\documentclass{article}

\usepackage[preprint]{neurips_2026}

\usepackage[utf8]{inputenc} 
\usepackage[T1]{fontenc}    
\usepackage{adjustbox}
\usepackage{algorithm}
\usepackage{algorithmic}
\usepackage{amsfonts}       
\usepackage{booktabs}       
\usepackage{float}
\usepackage{fvextra}
\usepackage{graphicx}       
\usepackage{hyperref}       
\usepackage{microtype}      
\usepackage{nicefrac}       
\usepackage{url}            
\usepackage{xcolor}         
\usepackage{wrapfig}
\usepackage{multirow}

\newcommand{\positive}[1]{{\scriptsize(\textcolor{green!50!black}{#1})}}
\newcommand{\negative}[1]{{\scriptsize(\textcolor{red!70!black}{#1})}}
\newcommand{\nofilter}{\textsc{NoFilter}}

\DefineVerbatimEnvironment{Verbatim}{Verbatim}{
  breaklines,
  breakanywhere,
  breaksymbolleft={},
  breaksymbolright={},
  breaksymbolsepleft=0pt,
  breaksymbolsepright=0pt,
  fontsize=\small,
}
\newcommand{\method} {Self-Verified Distillation}

\title{\method: Your Language Model Is Secretly Its Own Synthetic Data Pipeline}

\author{
  Tony Lee \\
  Stanford University \\
  \texttt{tonyhlee@stanford.edu}
  \And
  Percy Liang \\
  Stanford University \\
  \texttt{pliang@cs.stanford.edu}
}

\begin{document}

\maketitle

\begin{abstract}
Can post-trained large language models (LLMs) further improve themselves using only unlabeled prompts, without external teachers or feedback from tools? We study this setting starting only from unlabeled seed questions \textbf{with no ground-truth solutions}, across three reasoning domains: math, science, and coding. We propose \textbf{\method{}}, a simple post-training refinement algorithm in which the model generates candidate solutions to these seed questions, filters them using prompt-based self-verification, and trains on the resulting self-curated dataset. Inspired by the UQ benchmark's use of multiple validators to screen candidate answers to hard unsolved questions, we adapt this validation-based filtering idea to self-training: the model filters its own generated solutions through a three-stage cascade of cycle-consistency, factuality, and correctness checks, accepting a solution only if it passes all stages with unanimous judge votes.
We find that sampling more candidate generations and using a larger verification budget during training data construction produces higher-quality self-curated data and, in turn, better reasoning models.
We then train Qwen3 models at multiple scales with \method{} and obtain gains across all three domains.
For Qwen3-4B, our method improves aggregate held-out pass@1 by \textbf{+16.7} points in math (AIME26 and HMMT), \textbf{+11.1} points in science (GPQA Diamond and HLE), and \textbf{+8.3} points in coding (LCBv5 and LCBv6), with gains also extending to 0.6B and 8B models.
Compared to our test-time-only baseline (UQ-TTC), which improves performance by spending extra compute at inference time, \method{} achieves better performance in most settings while requiring only a single inference call at test time.
\end{abstract}

\begin{figure}[!h]
  \centering
  \vspace{-0.8em}
  \includegraphics[width=\linewidth]{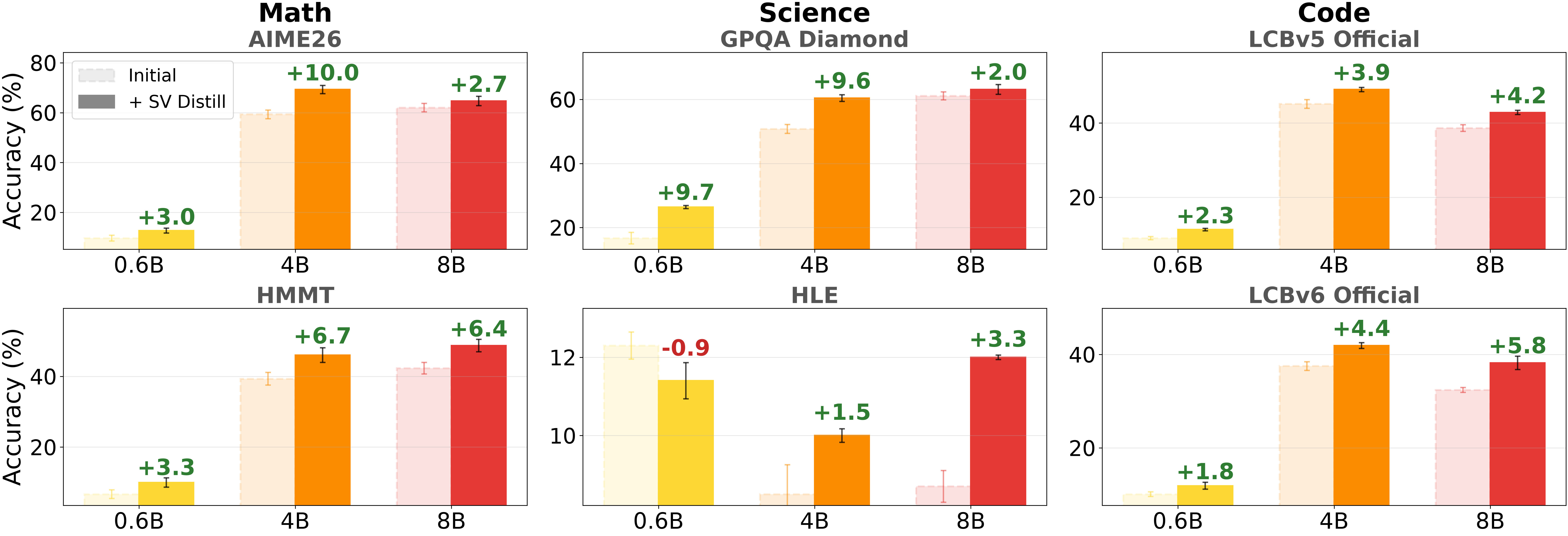}
  \vspace{-0.6em}
  \caption{\textbf{\method{} trains models on their own generations filtered by strong self-verification without ground-truth answers.}
  Across model scales, \method{} improves held-out pass@1 in math, science, and code, showing that self-verified training can further improve post-trained models. Error bars show $\pm$1 standard error of the mean across evaluation seeds.}
  \label{fig:main}
  \vspace{-0.95em}
\end{figure}

\newpage

\section{Introduction}

\begin{figure}[t]
  \centering
  \makebox[\linewidth][c]{%
    \includegraphics[width=1.023\linewidth]{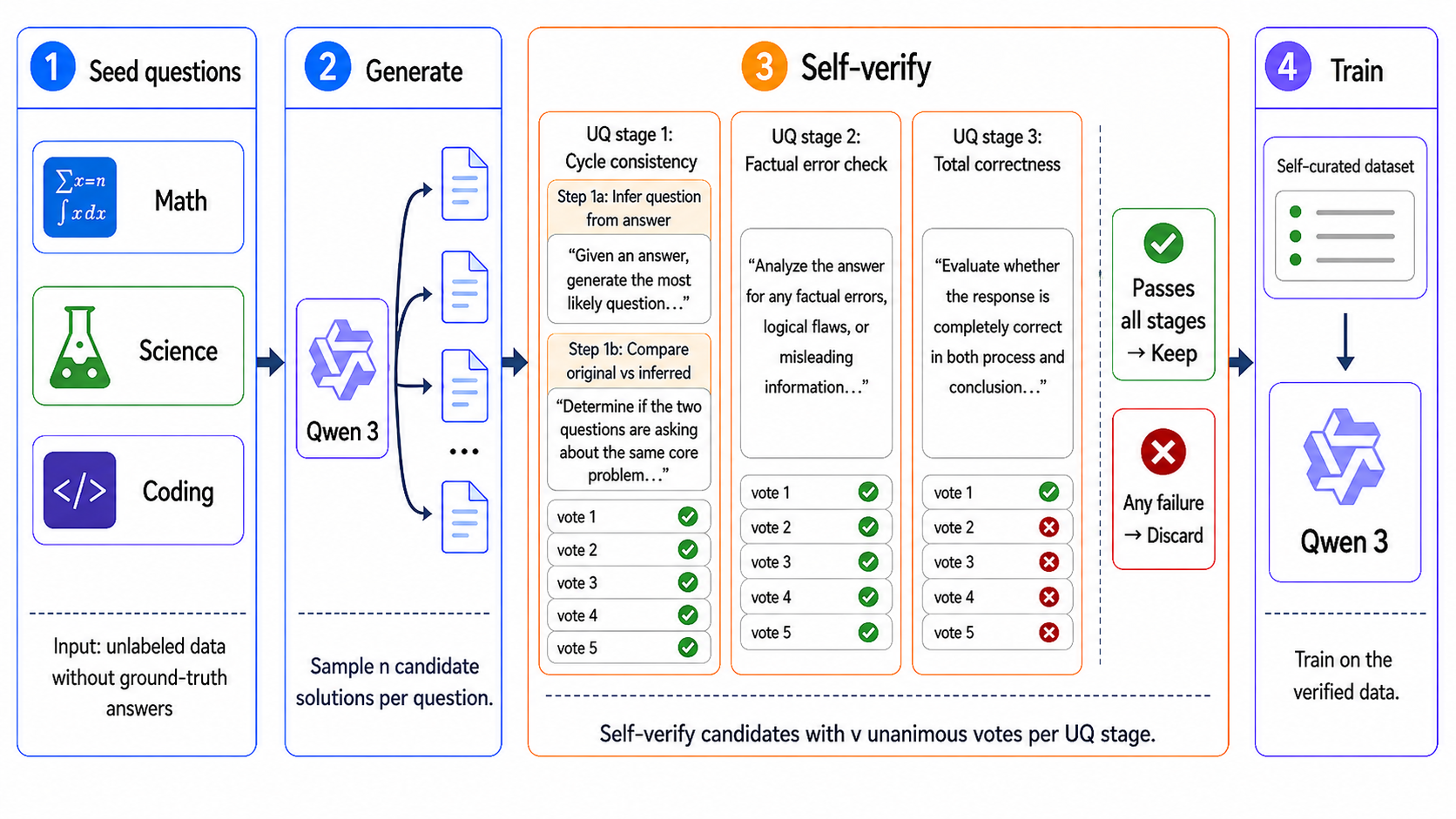}%
  }
  \vspace{-1.58em}
  \caption{Overview of \textbf{\method{}}. Starting from unlabeled seed questions without ground-truth answers, the model samples $n$ candidate solutions per question, filters each candidate with UQ-style multi-stage self-verification using $v$ repeated judge calls per stage, and trains on the accepted solutions as supervised data. This produces a self-curated dataset without external teachers, tool feedback, or ground-truth solutions.}
  \label{fig:overview}
\end{figure}

Despite rapid progress in post-training large language models for reasoning, further improving an already strong reasoning model remains challenging. Many successful approaches rely on additional sources of supervision, such as ground-truth answers for verifier training~\citep{cobbe2021trainingverifierssolvemath}, human-labeled process supervision~\citep{lightman2023lets}, or teacher-generated reasoning traces and synthetic answers~\citep{guha2025openthoughtsdatarecipesreasoning,abdin2025phi4reasoningtechnicalreport,zhu2026chimeracompactsyntheticdata}. These sources of supervision can be expensive to obtain, difficult to scale, or unavailable for the hardest reasoning problems. 
This issue arises across model scales and becomes especially important when stronger teachers are unavailable, inaccessible, or costly to use. This motivates a broader question for post-training: \emph{can an already post-trained reasoning model further improve using only itself?} 
This work builds on prior self-improvement and self-distillation approaches~\citep{huang2022largelanguagemodelsselfimprove,zelikman2022starbootstrappingreasoningreasoning,zhang2026embarrassinglysimpleselfdistillationimproves}, but studies whether explicit self-verification can make self-generated data reliable enough for further post-training in a deliberately constrained setting across math, science, and coding: \textbf{the model is given only unlabeled seed questions with no ground-truth solutions, no external teacher, and no tool feedback}. It must generate its own candidate solutions, decide which of those solutions are worth training on, and then improve from the resulting self-curated data.

The main challenge is whether a model can reliably verify its own generations. Naive self-training can reinforce the model's own mistakes: a model may generate incorrect solutions, judge them as valid, and then train on flawed reasoning traces. This challenge is closely related to generator-validator consistency, which studies whether a language model's behavior as a generator agrees with its behavior as a validator~\citep{li2023benchmarkingimprovinggeneratorvalidatorconsistency}. In our setting, self-verification is useful only if the model's validator view provides a higher-precision filter than the raw generator distribution: the verifier need not be perfect, but the accepted solutions must be better training data than unfiltered generations. To study this setting, we introduce \textbf{\method{}}, a simple post-training refinement algorithm that turns unlabeled seed questions into supervised training data. Starting from a post-trained model, \method{} samples $n$ candidate solutions for each seed question. It then filters these solutions using prompt-based, multi-stage self-verification inspired by the Unsolved Questions (UQ) benchmark, which uses compound validators to screen candidate answers to hard unsolved questions~\citep{nie2025uqassessinglanguagemodels}. Each candidate is checked by a cascade of cycle-consistency, factuality, and correctness verifiers, with $v$ repeated judge calls per stage, where a candidate is accepted only if it passes all checks. Finally, the model is trained on the accepted self-generated solutions (see Figure~\ref{fig:overview}).

A summary of our contributions is as follows:

\begin{enumerate}
    \item \textbf{Self-verified distillation from unlabeled questions.}
    We propose \textbf{\method{}}, a simple post-training refinement algorithm that further improves reasoning models starting only from unlabeled seed questions. Unlike many post-training pipelines, \method{} does not require external teachers, tool feedback, or ground-truth solutions: the model generates candidate solutions, self-verifies them, and trains on the resulting self-curated data.

    \item \textbf{A controlled study of self-curated data construction.}
    We study which parts of the synthetic data-generation pipeline matter by varying generation count, verification strength, verifier type, selection policy, and model scale across math, science, and coding. These experiments show that simply training on self-generated data can hurt performance, while stronger UQ verification improves the quality of the training data and outperforms a simpler prompt that checks correctness under the same verification strength.

    \item \textbf{Training-time verification versus test-time verification.}
    We introduce UQ-TTC, a test-time compute baseline that applies the same UQ-style verification procedure only at inference time. Compared to UQ-TTC, \method{} achieves competitive or stronger held-out math performance while requiring only a single inference call at evaluation time, showing that verification can be used to construct training data rather than repeatedly applied at test time.
\end{enumerate}

Figure~\ref{fig:main} gives an overview of the final held-out test results. Across Qwen3-0.6B, 4B, and 8B, \method{} improves pass@1 across math, science, and coding. The gains are strongest for Qwen3-4B, while the 0.6B results are smaller and less consistent, suggesting that model capability affects the quality of self-curated data. \method{} also performs competitively against our more computationally expensive UQ-TTC baseline on math, while using only a single inference call at evaluation time. These results suggest that \method{} can serve as a lightweight additional refinement stage for already post-trained reasoning models.

\section{Related Work}

\paragraph{Self-training.}
Our work builds on a long line of self-training and self-distillation methods. Some approaches bootstrap instruction-following data from model-generated instructions and outputs~\citep{wang2023selfinstructaligninglanguagemodels}, while others use self-generated rationale-augmented answers for self-improvement on unlabeled questions~\citep{huang2022largelanguagemodelsselfimprove}. Others train models to generate or use rationales more effectively. STaR iteratively bootstraps rationales by keeping those that lead to correct final answers~\citep{zelikman2022starbootstrappingreasoningreasoning}. Quiet-STaR instead learns to generate latent rationales that improve next-token prediction in general text~\citep{zelikman2024quietstarlanguagemodelsteach}. A related line of work uses model-generated critiques and revisions~\citep{bai2022constitutionalaiharmlessnessai} or multi-agent roles as self-improvement signals~\citep{chen2025multiagentevolvellmselfimprove}.

Recent work also studies related ways of obtaining learning signals from a model's own behavior. On-policy distillation trains on student-generated sequences while using a teacher to provide feedback on those sequences~\citep{agarwal2024onpolicydistillationlanguagemodels}. POPE uses privileged solution prefixes from humans or other models to guide on-policy exploration on hard reasoning problems~\citep{qu2026popelearningreasonhard}. Other work studies internal-feedback learning, where the model's own confidence provides a reward signal without external rewards or labeled data~\citep{zhao2026learningreasonexternalrewards}, or demonstration-conditioned self-distillation, where expert demonstrations are used in context to produce on-policy training signals~\citep{shenfeld2026selfdistillationenablescontinuallearning}. Self-Distillation Zero~\citep{he2026selfdistillationzeroselfrevisionturns}, a concurrent work, uses self-revision to turn binary correctness rewards into dense token-level supervision. In contrast, our setting does not assume access to ground-truth final answers, binary correctness rewards, expert demonstrations, external teachers, or tool feedback, and instead filters self-generated solutions using prompt-based multi-stage self-verification.

The closest prior work to our setting is Simple Self-Distillation (SSD), which improves code generation by directly training on a model's own raw, unverified outputs, without a stronger teacher, reward model, verifier, or labeled solutions~\citep{zhang2026embarrassinglysimpleselfdistillationimproves}. SSD's central finding is that this simple, unverified self-training can improve code generation. In contrast, our work studies math, science, and coding and focuses on the complementary question of whether self-generated data should be filtered before training. We show that simply training on a model's own generations can hurt performance, while applying strong self-verification to filter those generations yields further improvements across multiple domains.

\paragraph{Verification and rewards.}
Verification and reward models are central tools for improving reasoning models. Outcome verifiers are often trained to judge whether a generated solution is correct, and can be used to select among multiple sampled answers at inference time~\citep{cobbe2021trainingverifierssolvemath}. Process reward models instead score intermediate reasoning steps, providing finer-grained signals for ranking, reranking, or reinforcement learning over reasoning traces~\citep{lightman2023lets,wang2024mathshepherdverifyreinforcellms}. 

These verifiers and reward models are commonly used to filter, rank, or improve generated reasoning traces, either for test-time selection, rejection-sampling-style fine-tuning, reward-ranked fine-tuning, fine-grained reward-guided improvement, or verifier-guided search~\citep{dong2023raftrewardrankedfinetuning,zhang2024restmctsllmselftrainingprocess,hwang2024selfexploreenhancingmathematicalreasoning}. Recent work also studies how to combine multiple weak or imperfect verifiers into stronger aggregate verification systems~\citep{saadfalcon2025shrinkinggenerationverificationgapweak}. Relatedly, the Unsolved Questions (UQ) benchmark evaluates frontier models on hard questions without known ground-truth answers using compound, oracle-free validators~\citep{nie2025uqassessinglanguagemodels}. We adapt this validator-based filtering idea to select self-generated solutions for training.

\paragraph{Test-time compute.}
Our work is also related to test-time compute methods, which improve reasoning by spending more computation at inference time. One common approach is repeated sampling: generate many candidate solutions, then use a verifier, majority vote, unit tests, proof checkers, or reward model to select a final answer~\citep{brown2024largelanguagemonkeysscaling}. More generally, test-time compute can be viewed as modifying the model's prediction process at inference time, either by changing the input context or by sampling and selecting among multiple outputs~\citep{snell2024scalingllmtesttimecompute}. Other work explores larger inference-time systems that combine components such as generators, rankers, critics, verifiers, fusers, and unit-test evaluators under a fixed compute budget~\citep{saadfalcon2025archonarchitecturesearchframework}.

Recent work also studies self-improvement or aggregation entirely at test time. Some methods recursively generate, verify, and condition on previous attempts without updating model parameters~\citep{zhuang2026testtimerecursivethinkingselfimprovement}, while others maintain a population of candidate solutions and repeatedly aggregate or recombine them to produce stronger answers~\citep{venkatraman2026recursiveselfaggregationunlocksdeep}. Related approaches study distilled reasoners or societies of thought as ways to better use inference-time computation~\citep{paliotta2025thinkingslowfastscaling,kim2026reasoningmodelsgeneratesocieties}. In contrast, \method{} spends compute during data construction rather than only at test time: the model samples multiple candidate solutions, applies verification to filter candidates, and then amortizes the resulting compute into the model parameters through supervised fine-tuning.

Other work analyzes the structure and efficiency of reasoning traces themselves. Some studies find that reasoning models can overthink, producing unnecessarily long or repetitive reasoning traces on easier problems~\citep{chen2025think23overthinkingo1like}, while others show that models can underthink by abandoning promising lines of reasoning too early~\citep{wang2025thoughtsplaceunderthinkingo1like}. Related methods prefer shorter reasoning chains when they preserve correctness~\citep{hassid2026dontoverthinkitpreferring}, or explicitly explore alternative reasoning paths that were not taken in the original trajectory~\citep{lu2025retrosearchexploringuntakenpaths}. These works study how to spend or structure reasoning computation at inference time, while our work uses generation and verification compute to construct training data.

\section{\method}
\label{sec:self-instill}

\method{} is a post-training refinement pipeline with three components: (1) \textbf{generation}, where the model proposes candidate solutions for unlabeled seed questions, (2) \textbf{verification}, where prompt-based checks filter low-quality or incorrect responses, and (3) \textbf{training}, where the accepted solutions are used as supervised fine-tuning data. The resulting self-curated dataset allows the model to further train on its own verified generations. Figure~\ref{fig:overview} gives an overview of the pipeline, and Algorithm~\ref{alg:self_verified_distillation} provides the full procedure.

During generation, for each unlabeled seed question $q$, the initial model $p_{\theta}$ samples $n$ candidate solutions $\{y_1,\ldots,y_n\}$. These candidates are not assumed to be correct. They are treated as proposals that must undergo self-verification before being added to the training set. 
For verification, we are inspired by Unsolved Questions (UQ), a benchmark of challenging questions curated to lack known ground-truth answers, which uses automatic, oracle-free validators to pre-screen candidate answers for human review~\citep{nie2025uqassessinglanguagemodels}. Because such questions lack ground-truth answers, UQ cannot rely on standard answer matching for evaluation. Instead, it studies compound validation strategies that combine prompt-based checks, including cycle consistency, fact and logic checking, correctness checking, repeated or iterated judgments, and unanimous voting. We adapt this UQ-style verification from benchmark construction to post-training data construction: rather than using validators to decide whether an answer should enter an evaluation benchmark, we use them to decide whether a model-generated solution should be added to the supervised training dataset. Each candidate solution is checked by a cascade of cycle-consistency, factuality, and correctness verifiers, and is accepted only if it passes all stages. Full verifier prompts are provided in Appendix~\ref{app:uq_self_verification}.

We use two data-construction parameters: $n$, the number of candidate solutions sampled per seed question, and $v$, the number of repeated judge calls per verification stage. Increasing $n$ gives the model more opportunities to produce a useful solution for each seed question. Increasing $v$ makes verification more conservative, since \textbf{a candidate solution must pass every repeated judge call at every stage}. Therefore, during data construction, $n$ controls exploration over candidate solutions, while $v$ controls verification stringency. Algorithm~\ref{alg:self_verified_distillation} summarizes the full \method{} pipeline.

\begin{algorithm}[t]
\caption{\textbf{\method{}}. Given unlabeled seed questions, the model samples $n$ candidate solutions per question, applies three-stage self-verification with $v$ repeated judge calls per stage, and trains on the accepted solutions with supervised fine-tuning.}
\label{alg:self_verified_distillation}
\begin{algorithmic}[1]
\REQUIRE
Initial post-trained model $p_{\theta}$ \\
Unlabeled seed questions $\mathcal{Q}$ \\
Number of generations $n$ \\
Verification strength $v$ \textnormal{(number of repeated judge calls per verification stage)}
\ENSURE
Refined model $p_{\theta'}$

\STATE Initialize self-curated dataset $\mathcal{D}_{\mathrm{sft}} \gets \emptyset$

\FOR{each seed question $q \in \mathcal{Q}$}
    \STATE Generate candidate solutions $\{y_1,\ldots,y_n\} \sim p_{\theta}(\cdot \mid q)$

    \FOR{each candidate solution $y_i$}
        \STATE Evaluate self-verifier $V_{\theta,\mathrm{cycle}}(q,y_i)$ for $v$ repeated judge calls
        \STATE Evaluate self-verifier $V_{\theta,\mathrm{fact}}(q,y_i)$ for $v$ repeated judge calls
        \STATE Evaluate self-verifier $V_{\theta,\mathrm{corr}}(q,y_i)$ for $v$ repeated judge calls
        
        \IF{$y_i$ passes all three verifier stages in all repeated judge calls}
            \STATE Add $(q,y_i)$ to $\mathcal{D}_{\mathrm{sft}}$
        \ENDIF
    \ENDFOR
\ENDFOR

\STATE Train $p_{\theta}$ on $\mathcal{D}_{\mathrm{sft}}$ with supervised fine-tuning:
$\theta' \gets \mathrm{SFT}(\theta, \mathcal{D}_{\mathrm{sft}})$

\STATE \textbf{return} $p_{\theta'}$
\end{algorithmic}
\end{algorithm}

\section{Experiments}
 
Our experiments are organized to isolate the main design choices in the self-training synthetic data pipeline before evaluating whether the resulting recipe transfers across model scales. We first study two core uses of data-construction compute: $n$, the number of candidate generations sampled per seed question, and $v$, the verification strength used to filter those generations. We then ablate the type of verifier and compare selection policies for converting accepted candidates into training data. Finally, we evaluate the resulting recipe across Qwen3 models of different sizes and compare using verification for training data construction against using the same verifier only at test time.

\subsection{Setup}
\label{sec:setup}

We use a shared setup across experiments, varying only the data-construction component under study. Starting from a set of unlabeled seed questions, we generate candidate solutions with fixed inference hyperparameters, filter outputs, train with a fixed SFT recipe, and evaluate on validation and held-out test benchmarks across math, science, and coding. This setup allows differences across ablations to primarily reflect changes in the self-curated training data rather than changes in prompting, optimization, or evaluation.

\paragraph{Seed questions} We initialize our starting question set from OpenThoughts~\citep{guha2025openthoughtsdatarecipesreasoning}: the math subset (53,125 examples), and expanded subsets for science (26,041 examples) and code (9,168 examples). We use only the questions from OpenThoughts and disregard the provided reasoning traces and answers to remain consistent with settings where the objective is to improve reasoning without ground-truth solutions. Additional details and representative seed questions are provided in Appendix~\ref{app:seed_questions}.


\paragraph{Inference} We use the same inference hyperparameters for both generation and verification: temperature of 0.8, top-p of 0.95, and a maximum output length of 32,768 tokens, the recommended output length for Qwen3 models. During data generation, we immediately filter out malformed generations (e.g., incomplete reasoning traces or missing final answers).

\paragraph{Training} We use a fixed training recipe across ablations so that differences between runs primarily reflect differences in the generated training data rather than optimization choices. Training hyperparameters and compute details are provided in Appendix~\ref{app:training}.

\paragraph{Evaluation} We evaluate across math, science, and coding benchmarks. We report pass@1, averaged across random seeds for benchmarks evaluated with multiple seeds. For math, we use AIME24 (x10 random seeds), AIME25 (x10), MATH500, and OlympiadBench (English-only math subset) for validation, and AIME26 (x10) and HMMT 02/25 (x10) for test. For science, we use JEEBench (x3) and OlympiadBench (English-only physics subset) for validation, and GPQA Diamond (x3) and HLE (x3) for test. For coding, we use LiveCodeBench v2 (5/2023--5/2024, x6) for validation, and LiveCodeBench v5 Official (8/2024--1/2025, x6) and LiveCodeBench v6 Official (1/2025--4/2025, x6) for test. We select checkpoints based on validation performance and report final results on the held-out test benchmarks. Additional evaluation details are provided in Appendix~\ref{app:evaluations}.

\subsection{Number of Generations and Verification Strength}

\paragraph{Self-verified distillation benefits from both broader generation and stronger verification.}
We study how \method{} depends on the two main data-construction choices: the number of candidate generations $n$ and the verification strength $v$. This experiment tests whether greater self-improvement comes from broader exploration, more selective filtering, or both. Increasing $n$ gives the model more opportunities to produce a useful solution for each seed question, while increasing $v$ makes the verifier more conservative by requiring repeated judge calls to agree at each verification stage. We conduct controlled ablations of \method{} across three domains (math, science, and coding) using the same post-trained Qwen3-4B model~\citep{qwen3technicalreport}.

\begin{figure}[t]
  \centering
    \includegraphics[width=\textwidth]{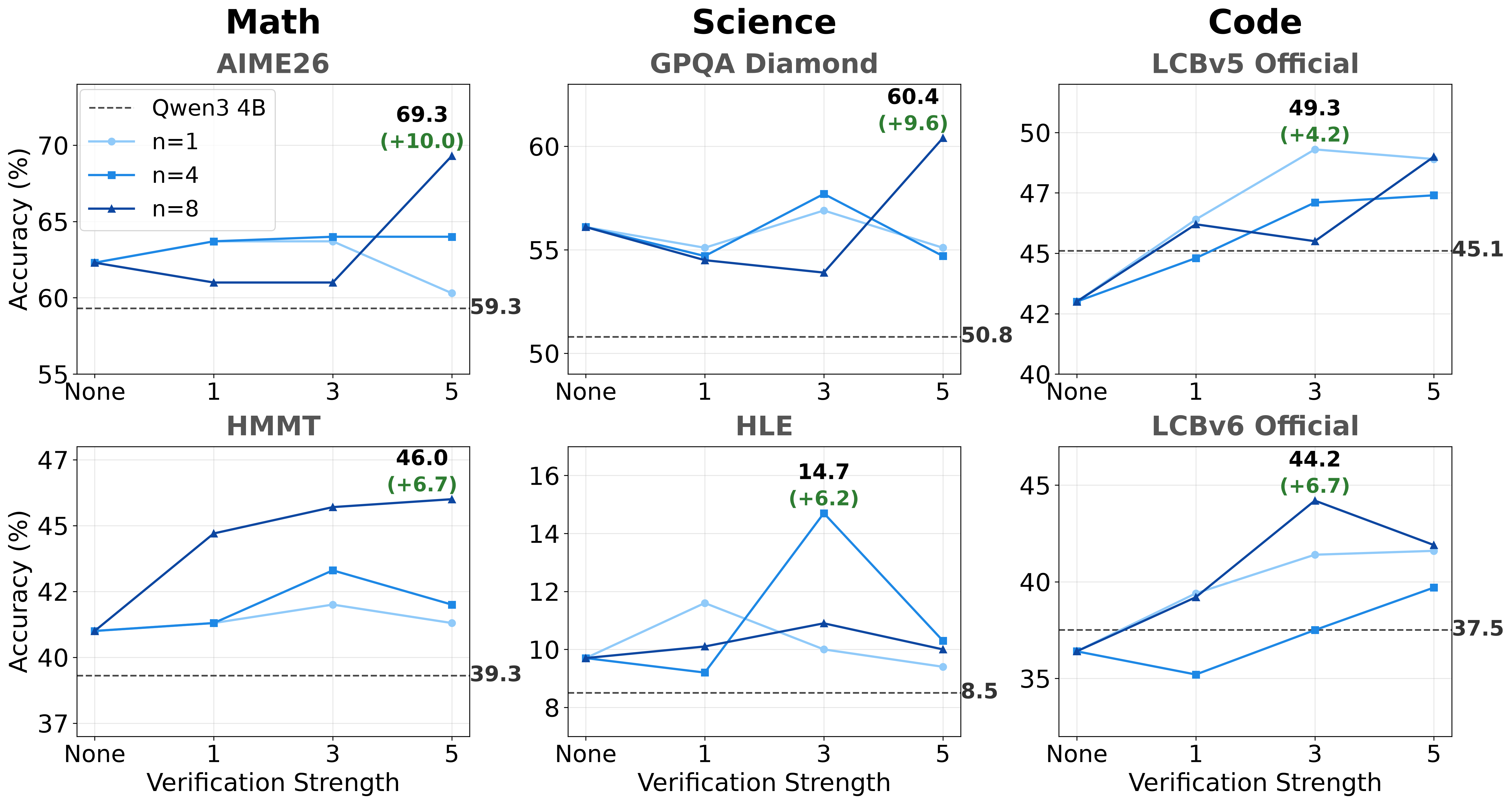}
    \caption{Effect of generation count and verification strength for Qwen3-4B. We vary the number of candidate generations $n \in \{1,4,8\}$ and the verification strength $v \in \{\nofilter,1,3,5\}$ across math, science, and coding held-out test benchmarks. Here, $v=\nofilter$ denotes training on unfiltered generations, and the dashed horizontal line shows the initial Qwen3-4B model. Using more generation and verification during data construction generally improves performance.}
    \label{fig:num_generations_filter_strength}
\end{figure}

Figure~\ref{fig:num_generations_filter_strength} shows that increasing compute for generation and verification during data construction generally improves downstream performance, but the allocation of compute between generation and verification matters. In math, the largest generation and verification setting, $n=8$ and $v=5$, gives the best held-out results on both AIME26 and HMMT, improving over the initial model by +10.0 and +6.7 points, respectively. This suggests that, for this seed-question distribution and evaluation setup, combining more candidate generations with stronger verification produces the best self-curated training data. However, this trend is not simply monotonic in verification strength alone: when $n$ is small, increasing $v$ can make the filter too conservative and leave too few accepted solutions for training. Strong verification is most useful when paired with enough candidate generations, since a larger candidate pool can better absorb the lower acceptance rate.

Code provides the clearest evidence that filtering is necessary. Training on unfiltered generations performs worse than the initial model on both LCBv5 and LCBv6, while verified data improves performance. \textbf{This shows that simply training on unfiltered generations can hurt performance, while filtering those generations with self-verification can turn the same self-generated data into a useful training signal.} The best code settings improve LCBv5 by +4.2 points and LCBv6 by +6.7 points. Science shows the same general trend, but with more variation across benchmarks: $n=8$, $v=5$ gives the best GPQA Diamond result, improving by +9.6 points, while HLE performs best with $n=4$, $v=3$, improving by +6.2 points. Overall, these results show that simply generating more data is not sufficient and that the generated data must also be filtered carefully. They also highlight an interaction between exploration and verification: stronger filtering can improve quality, but only if enough generations are sampled to preserve sufficient training coverage. 
This motivates studying how $n$ and $v$ jointly shape the quality and coverage of the self-curated training data.

\subsection{Type of Verification}
\label{sec:type_of_verification}

\paragraph{The structure of verification matters, not just the verification strength.}
The previous experiment varies verification strength, but not the structure of the verifier itself. We next ask whether stronger results come only from more stringent verification, or whether the form of verification also matters. This distinction is important because a simple correctness prompt is easier to implement in practice, while the UQ-style verifier decomposes verification into multiple checks that probe different failure modes.

We use the same setting as the Qwen3-4B math ablation above: for each seed question, we sample $n=8$ candidate solutions, apply verification with strength $v=5$, and then train on the accepted solutions. The only difference between the two variants is the verification filter. The simple verifier directly asks whether each generated solution is correct or incorrect. The UQ-style verifier instead uses the full multi-stage cascade of cycle-consistency, factuality, and correctness checks. The full prompt for the simple verifier is provided in Appendix~\ref{app:alternative_verification}.

\begin{wraptable}{r}{0.48\linewidth}
\vspace{-2em}
\caption{Effect of verification type on held-out math benchmarks. Mean $\Delta$ is the average improvement over the initial model.}
\label{tab:verification_type}
\centering
\small
\begin{tabular}{lccc}
\toprule
Verification & AIME26 & HMMT & Mean $\Delta$ \\
\midrule
Simple & 64.7 \positive{+5.4} & 43.7 \positive{+4.4} & +4.9 \\
Full UQ & \textbf{69.3} \positive{+10.0} & \textbf{46.0} \positive{+6.7} & \textbf{+8.4} \\
\bottomrule
\end{tabular}
\vspace{-2em}
\end{wraptable}

As shown in Table~\ref{tab:verification_type}, the UQ-style multi-stage verifier substantially outperforms the simple verifier under the same generation count, verification strength, and training setup. The simple verifier already improves over the initial model, suggesting that even direct prompt-based filtering is useful. However, replacing it with the UQ-style verifier increases the mean held-out improvement from $+4.9$ to $+8.4$ points. This indicates that decomposing verification into cycle-consistency, factuality, and correctness checks produces self-curated data that leads to stronger reasoning models compared to using the simple verifier.

\subsection{Effect of Model Scale}
\label{sec:model_scale}

\begin{table}[t]
    \caption{Effect of model scale. We apply \method{} with $n=8$ candidate generations and verification strength $v=5$ to Qwen3 models of different sizes and report held-out test performance across math, science, and coding. Values in parentheses show absolute improvement over the corresponding initial model. Mean $\Delta$ is the average improvement across the two held-out benchmarks in each domain.}
    \label{tab:self_instilled_model_scale}
    \centering
    \resizebox{\linewidth}{!}{%
    \begin{tabular}{lccccccccc}
    \toprule
    \textbf{Models}
    & \multicolumn{3}{c}{\textbf{Math}}
    & \multicolumn{3}{c}{\textbf{Science}}
    & \multicolumn{3}{c}{\textbf{Code}} \\
    \cmidrule(lr){2-4} \cmidrule(lr){5-7} \cmidrule(lr){8-10}
    & \textbf{AIME26} & \textbf{HMMT} & \textbf{Mean $\Delta$}
    & \textbf{GPQA Diamond} & \textbf{HLE} & \textbf{Mean $\Delta$}
    & \textbf{LCBv5} & \textbf{LCBv6} & \textbf{Mean $\Delta$} \\
    \midrule
    Qwen3-0.6B & 9.7 & 6.7 & -- & 16.7 & 12.3 & -- & 9.0 & 10.1 & -- \\
    SV Distillation & \textbf{12.7} \positive{+3.0} & \textbf{10.0} \positive{+3.3} & \textbf{+3.2} & \textbf{26.4} \positive{+9.7} & 11.4 \negative{-0.9} & \textbf{+4.4} & \textbf{11.3} \positive{+2.3} & \textbf{11.9} \positive{+1.8} & \textbf{+2.1} \\
    \midrule
    Qwen3-4B & 59.3 & 39.3 & -- & 50.8 & 8.5 & -- & 45.1 & 37.5 & -- \\
    SV Distillation & \textbf{69.3} \positive{+10.0} & \textbf{46.0} \positive{+6.7} & \textbf{+8.4} & \textbf{60.4} \positive{+9.6} & \textbf{10.0} \positive{+1.5} & \textbf{+5.6} & \textbf{49.0} \positive{+3.9} & \textbf{41.9} \positive{+4.4} & \textbf{+4.2} \\
    \midrule
    Qwen3-8B & 62.0 & 42.3 & -- & 61.1 & 8.7 & -- & 38.6 & 32.4 & -- \\
    SV Distillation & \textbf{64.7} \positive{+2.7} & \textbf{48.7} \positive{+6.4} & \textbf{+4.6} & \textbf{63.1} \positive{+2.0} & \textbf{12.0} \positive{+3.3} & \textbf{+2.7} & \textbf{42.8} \positive{+4.2} & \textbf{38.2} \positive{+5.8} & \textbf{+5.0} \\
    \bottomrule
    \end{tabular}%
    }
\end{table}

\paragraph{\method{} improves performance across Qwen3 model scales.} 
We next evaluate whether the same data-construction recipe transfers across model scales. Since \method{} depends on both generation and verification, model capability may affect the quality of the self-curated data. 
We therefore apply the same \method{} recipe, using $n=8$ candidate generations and verification strength $v=5$, across Qwen3-0.6B, Qwen3-4B, and Qwen3-8B, with the same held-out test benchmarks across math, science, and coding.

Table~\ref{tab:self_instilled_model_scale} shows that \method{} improves mean held-out performance across all three model sizes and all three domains. For Qwen3-0.6B, \method{} improves mean performance by +3.2 points in math, +4.4 points in science, and +2.1 points in coding. The gains for this weakest model are smaller and less consistent: in particular, performance on HLE decreases slightly from 12.3 to 11.4. For Qwen3-4B, the gains are stronger, with mean improvements of +8.4 points in math, +5.6 points in science, and +4.2 points in coding. For Qwen3-8B, \method{} continues to improve performance, with mean gains of +4.6 points in math, +2.7 points in science, and +5.0 points in coding.

These results suggest that \method{} is not limited to a single model scale. At the same time, the gains are not monotonic in model size, and the weakest model shows no improvement on one benchmark (HLE). One possible explanation is that a fixed seed-question distribution provides different amounts of useful learning signal for different initial models: weaker models may struggle to generate or verify high-quality solutions, while stronger models may already solve many of the easier seed questions. More broadly, these results suggest that the difficulty and diversity of seed questions may need to be matched to the capability of the initial model, which we leave to future work.

\subsection{Training vs. Test-Time Compute}
\label{sec:training_vs_ttc}

\paragraph{\method{} is stronger than test-time verification in most settings, while using a single inference call at evaluation time.}
A natural alternative to \method{} is to use the same self-verification procedure only at test time. In this setting, the model samples multiple candidate solutions for each test problem, applies the UQ-style verifier to those candidates, and returns an accepted solution. This directly improves inference, but it is expensive because the generation and verification costs are paid for every test example. In contrast, \method{} pays this cost upfront during training data construction, trains on the accepted solutions, and then evaluates the trained model with a single generation per test example.

We compare these two ways of using the same verification strength on held-out math benchmarks. For test-time compute with UQ verification (UQ-TTC), we use the same $n=8$, $v=5$ setting as our best math SFT recipe, applying the prompts from Appendix~\ref{app:uq_self_verification} at inference time. This requires sampling up to 8 candidate solutions and running repeated verifier calls for each candidate. In our UQ-style verifier, cycle-consistency requires two inference calls, one to infer the question from the candidate solution and one to judge whether the inferred question matches the original question, followed by fact/logic and correctness checks. Thus, UQ-TTC can require up to $8$ candidate generations plus $8 \times 4 \times 5 = 160$ verifier calls, for up to $168$ total inference calls per test problem. By contrast, after \method{} training, evaluation uses only a single model generation per test problem.

\begin{table}[t]
\caption{Training on verified data versus using UQ verification only at test time. We compare the initial Qwen3 models of various sizes, \method{} training with the best math recipe, and UQ test-time compute (UQ-TTC) using the same $n=8$, $v=5$ verification setting. Max inference calls denotes the maximum total number of inference calls at evaluation time \emph{per test problem}, including candidate generations and verifier calls. Values in parentheses show absolute improvement over the corresponding initial model.}
\label{tab:training_vs_ttc}
\centering
\small
\begin{tabular}{llccc}
\toprule
Model size & Method & Max inference calls & AIME26 & HMMT \\
\midrule
\multirow{3}{*}{0.6B}
& Initial model & 1 & 9.7 & 6.7 \\
& \method{} & 1 & \textbf{12.7} \positive{+3.0} & \textbf{10.0} \positive{+3.3} \\
& UQ-TTC & 168 & 10.7 \positive{+1.0} & 9.3 \positive{+2.6} \\
\midrule

\multirow{3}{*}{4B}
& Initial model & 1 & 59.3 & 39.3 \\
& \method{} & 1 & \textbf{69.3} \positive{+10.0} & \textbf{46.0} \positive{+6.7} \\
& UQ-TTC & 168 & 68.0 \positive{+8.7} & 45.3 \positive{+6.0} \\
\midrule

\multirow{3}{*}{8B}
& Initial model & 1 & 62.0 & 42.3 \\
& \method{} & 1 & 64.7 \positive{+2.7} & \textbf{48.7} \positive{+6.4} \\
& UQ-TTC & 168 & \textbf{67.7} \positive{+5.7} & 47.0 \positive{+4.7} \\
\bottomrule
\end{tabular}
\end{table}

Table~\ref{tab:training_vs_ttc} shows that both uses of verification improve over the initial models on every held-out benchmark. Compared to UQ-TTC, \method{} performs better on five of the six individual benchmark comparisons while using only a single inference call at evaluation time. For Qwen3-0.6B and Qwen3-4B, \method{} outperforms UQ-TTC on both AIME26 and HMMT. For Qwen3-8B, UQ-TTC is stronger on AIME26, while \method{} is stronger on HMMT. Overall, these results suggest that training and test-time verification are complementary ways to use the same verification procedure. UQ-TTC can give strong per-example gains, but \textbf{\method{} amortizes the data-construction compute into the model, yielding pass@1 improvements without requiring multi-sample generation and repeated verification at evaluation time}.

\section{Discussion}
\label{sec:discussion}

In this work, we introduced \textbf{\method{}}, a simple post-training refinement method that turns unlabeled seed questions into supervised training data using the model's own generations and prompt-based self-verification. Across math, science, and coding, we find that post-trained Qwen3 models can further improve without external teachers, tool feedback, or ground-truth solutions. Our ablations show that these gains depend on the data-construction pipeline: unfiltered generations can hurt performance, stronger verification can improve the quality of the self-curated data, and UQ-style multi-stage verification outperforms a simpler correctness prompt under the same verification strength. 
More broadly, our results suggest that generation and verification compute can be useful not only at test time, but also during training-data construction: \method{} samples multiple candidate solutions, filters them with strong self-verification, and amortizes this compute into the model through supervised fine-tuning.

At the same time, \method{} has important limitations and risks. Self-verification is imperfect: even with a conservative multi-stage verifier, the model may accept incorrect solutions, reject useful ones, or apply verification unevenly across domains. The method also depends on the difficulty and diversity of the seed questions. A fixed seed distribution may provide different amounts of useful learning signal for different model scales, suggesting that future work should study how to match seed-question difficulty to the capability of the initial model. More broadly, training on self-generated data can reinforce systematic mistakes, overfit to the verifier's preferences, or make models more confident in flawed reasoning patterns. Future work should explore alternative verification methods, better mitigation strategies, and richer seed-question distributions.


\bigskip
\subsection*{Acknowledgments}

We thank Moo Jin Kim, Kaiyue Wen, Ken Liu, Rohith Kuditipudi, and Suhas Kotha for helpful discussions related to this work. This work is part of the \href{https://marin.community/}{Marin Project}, and the compute was supported by the Google TPU Research Cloud (TRC) program. This research was supported by Schmidt Sciences.



\bibliographystyle{plainnat}
\bibliography{references}


\newpage

\appendix

\section{Additional Related Work}
\label{app:additional_related_work}

\paragraph{Synthetic reasoning data for training.}
Recent reasoning models are often improved through synthetic data, teacher-generated reasoning traces, rejection sampling, reinforcement learning, data filtering, and carefully tuned post-training recipes. DeepSeek-R1 demonstrates that reinforcement learning can incentivize reasoning behavior in large language models~\citep{Guo_2025}, while Qwen3 uses large-scale pretraining and synthetic math and code data as part of its reasoning-oriented training pipeline~\citep{qwen3technicalreport}. OpenThoughts studies data recipes for reasoning models, including source-question selection, teacher answer generation, filtering, and evaluation across math, science, and code~\citep{guha2025openthoughtsdatarecipesreasoning}. Phi-4-reasoning trains on large-scale prompts with teacher-generated long reasoning traces and then applies a further reinforcement-learning stage for the ``plus'' model~\citep{abdin2025phi4reasoningtechnicalreport}. CHIMERA constructs a compact synthetic reasoning dataset using state-of-the-art models for problem generation, verification, and solution synthesis before post-training a Qwen3 model~\citep{zhu2026chimeracompactsyntheticdata}. These approaches show the importance of high-quality synthetic reasoning data, but they often rely on stronger teacher models, generated answers with known correctness, external verifiers, or model-specific data construction pipelines.

Other work studies how to target or reshape reasoning supervision. Skill-targeted adaptive training generates synthetic data for missing skills~\citep{he2025skilltargetedadaptivetraining}, long-chain reasoning distillation studies adaptive alignment for transferring long reasoning behavior~\citep{liu2026longchainreasoningdistillationadaptive}, and recent work examines when reinforcement learning helps reasoning in smaller LLMs~\citep{dang2026reinforcementlearningreasoningsmall}. Related methods improve small models with self-evolved reasoning~\citep{guan2025rstarmathsmallllmsmaster}. There is also work on making reasoning supervision more efficient by training on compressed chain-of-thought traces~\citep{kang2024c3otgeneratingshorterchainofthought} or by augmenting data with inferred latent thoughts~\citep{ruan2025reasoninglearnlatentthoughts}. Our setting is more constrained: we start only from unlabeled seed questions, discard any provided answers or reasoning traces, and ask whether the same post-trained model can generate, verify, and learn from its own candidate solutions without external teachers, tool feedback, or ground-truth solutions.

\section{Self-Verification}
\label{app:uq_self_verification}

We use a multi-stage self-verification pipeline inspired by the validation procedure in Unsolved Questions (UQ)~\citep{nie2025uqassessinglanguagemodels}. UQ studies hard unsolved questions, where ground-truth answers are unavailable, and therefore uses automatic, oracle-free validators to screen candidate answers. We adapt this idea from benchmark construction to post-training data construction: instead of using validators to decide whether a candidate answer should enter an evaluation benchmark, we use them to decide whether a model-generated solution should enter the supervised fine-tuning dataset.

Our verifier has three stages. First, \textbf{cycle consistency} checks whether the generated solution appears to answer the original question by inferring the question from the answer and comparing it to the seed question. Second, \textbf{factual error checking} screens for factual, arithmetic, or logical errors in the response. Third, \textbf{total correctness} checks whether the answer is complete and solves the original question. For verification strength $v$, each stage is repeated $v$ times, and a candidate solution is accepted only if it passes every stage in all repeated judge calls. 

\subsection{Cycle Consistency}

Cycle consistency consists of two prompts. The first prompt infers the question that the answer appears to address. The second prompt compares the inferred question to the original seed question.

\paragraph{Question generation prompt.}

\begin{Verbatim}
Given an answer, please generate the most likely question that would have prompted this answer. Focus on inferring the core question that this answer is addressing. Output only the inferred question, without any additional explanation.

Answer:
{answer}

Inferred Question:
\end{Verbatim}

\paragraph{Question comparison prompt.}

\begin{Verbatim}
You are evaluating whether an answer is relevant to the original question and touches the core of the question by comparing the original question with an inferred question derived only from the answer.

Original Question: {original_question}
Inferred Question: {inferred_question}

Compare the two questions and determine:
1. If the original question and inferred question are asking about the same core topic
2. If the original question and inferred question share the same key elements and requirements
3. If answering one question would effectively address the other question

After your analysis, provide your decision: [[Y]] if the questions are semantically equivalent and address the same core problem, or [[N]] if they are asking about different things.
\end{Verbatim}

\subsection{Factual Error Check}

The factual error check screens whether the candidate solution contains factual, arithmetic, or logical errors that would make it unsuitable for training.

\begin{Verbatim}
Please act as an impartial judge and carefully analyze the following answer for any factual errors, logical flaws, or misleading information.

Question: {question}
Answer: {answer}

Consider the credibility of the claims made in the answer and determine if they align with established knowledge. Evaluate:
1. Are there any incorrect facts, dates, numbers, formulas, or claims?
2. Is there any faulty logic, reasoning, or problem-solving approach?
3. Are there any misleading, incomplete, or ambiguous explanations?
4. Does the answer introduce any misconceptions or propagate common errors?

Minor typos or grammatical errors are acceptable. But be strict about any factual error, calculation error, or logical flaw. When unsure, lean toward accepting statements unless they contain clear errors.

After a thorough analysis, provide your decision: [[Y]] if the answer has no factual errors or major flaws, or [[N]] if it contains important factual errors or logical flaws that would mislead the user.
\end{Verbatim}

\subsection{Total Correctness}

The total correctness check evaluates whether the candidate solution fully solves the seed question. This stage is intentionally strict: partial, incomplete, or high-level answers are rejected.

\begin{Verbatim}
Please act as an impartial judge and evaluate whether the response is completely correct in both process and conclusion.

Question: {question}
Answer: {answer}

Consider correctness, usefulness, completeness and depth in your assessment. Consider whether this answer completely solves the question.

You should rely on your own reasoning to form a reference solution and compare the answer to your reasoning.

Begin your evaluation by giving a brief summary of your thoughts on the response. Focus on whether it is accurate, addresses the question well, and is reasonably detailed. Be precise about any errors or gaps you notice.

Notes:
1. If the answer is partial, high-level, or just states that this is an open problem, you should not accept it.
2. If the answer lacks details or is not comprehensive, you should not accept it.
3. If the answer contains any errors, you should not accept it.
4. You should only accept the answer if it is at least 95% correct and solves the question.

After providing your explanation, decide whether this answer is correct. Think twice about whether this answer solves the question.
Format: Accepted: [[Y]] if you accept the answer or Accepted: [[N]] if you do not accept.
\end{Verbatim}

\section{Experimental Details}

\subsection{Setup Details}

\subsubsection{Seed Questions}
\label{app:seed_questions}

We use OpenThoughts~\citep{guha2025openthoughtsdatarecipesreasoning} as the source of unlabeled seed questions. OpenThoughts studies data recipes for reasoning models across math, science, and coding, and its datasets include questions together with model-generated reasoning traces and answers. In our setting, we use only the question portion of each example and discard any provided answers or reasoning traces. This matches our no-ground-truth setting: the model must generate candidate solutions from the seed questions, self-verify those candidates, and train only on the accepted self-generated data.

Concretely, we use 53,125 math seed questions, 26,041 science seed questions, and 9,168 code seed questions from OpenThoughts-derived subsets. The seed questions span a range of formats and difficulty levels. Math seeds include competition-style combinatorics, number theory, geometry, algebra, and graph problems. Science seeds include chemistry, physics, and conceptual scientific reasoning questions, ranging from short multiple-choice items to longer open-ended questions. Code seeds include algorithmic programming problems, code-golf-style problems, and input-output programming tasks with constraints.

Table~\ref{tab:seed_question_examples} gives representative examples of seed questions from each domain. These examples illustrate the diversity of the starting questions; no ground-truth solutions from these examples are used during \method{}. The seed questions are used only as prompts for data construction. For each seed question $q$, the initial model samples $n$ candidate solutions, applies the self-verification pipeline described in Appendix~\ref{app:uq_self_verification}, and adds only accepted solutions to the supervised fine-tuning dataset. Thus, the final training data is determined by both the seed-question distribution and the model's ability to generate and verify useful solutions.

\begin{table}[t]
\caption{Representative unlabeled seed questions used for self-verified distillation. We use only the question text from OpenThoughts-derived examples and discard provided answers or reasoning traces.}
\label{tab:seed_question_examples}
\centering
\small
\begin{adjustbox}{max width=\linewidth}
\begin{tabular}{lp{0.78\linewidth}}
\toprule
Domain & Example seed question \\
\midrule
Math &
Five people, Alex, Bob, Charlie, Dave, and Emily, are randomly seated around a circular table. What is the probability that no two people with consecutive names are seated next to each other? \\
\addlinespace
Math &
Determine the largest positive integer $x$ for which there exists a unique positive integer $n$ such that $x$ can be expressed as a sum of $n$ distinct perfect cubes in at least two different ways. \\
\addlinespace
Science &
Which of the following are correct statements about white phosphorus ($P_4$)? (a) Each P atom has one lone pair of electrons in a molecule of $P_4$; (b) it has tetrahedral structure with 109$^\circ$28$'$ bond angle; (c) there are six P--P linkages in a molecule of $P_4$; (d) it undergoes disproportionation on reaction with alkalies. \\
\addlinespace
Science &
Compound (A), $\mathrm{C}_5\mathrm{H}_{10}\mathrm{O}$, forms a phenylhydrazone, gives negative Tollens' and iodoform tests, and is reduced to pentane. What is the compound? \\
\addlinespace
Code &
Given the infinite sequence formed by concatenating $1, 12, 123, 1234, \ldots$, return the digit at index $n$, where $1 \leq n \leq 10^{18}$. \\
\addlinespace
Code &
There are $n$ computers in a row, initially off. Phoenix manually turns on computers one at a time, while some computers may turn on automatically when both neighbors are on. Count the number of possible manual turn-on sequences modulo a prime $M$. \\
\bottomrule
\end{tabular}
\end{adjustbox}
\end{table}

\subsubsection{Training}
\label{app:training}

We train all models with supervised fine-tuning on the self-curated datasets produced by \method{}. Table~\ref{tab:training_hyperparams} summarizes the training hyperparameters for the final $n=8$, $v=5$ runs across domains and model scales. Batch size denotes the effective batch size, with the micro-batch or gradient-accumulation configuration shown in parentheses.

\begin{table}[H]
\caption{Training hyperparameters.}
\label{tab:training_hyperparams}
\centering
\small
\begin{adjustbox}{max width=\linewidth}
\begin{tabular}{llcccccc}
\toprule
Domain & Scale & Steps & LR & Warmup & WD & Clip & Batch \\
\midrule
Math & 0.6B & 2,000 & $3\mathrm{e}{-5}$ & 0.05 & 0.01 & 1.0 & 64 (64) \\
Math & 4B & 4,000 & $2\mathrm{e}{-5}$ & 0.03 & 0.0 & 0.2 & 64 (64) \\
Math & 8B & 2,000 & $5\mathrm{e}{-6}$ & 0.05 & 0.01 & 1.0 & 64 (32 $\times$ 2 accum.) \\
\midrule
Code & 0.6B & 2,000 & $3\mathrm{e}{-5}$ & 0.05 & 0.01 & 1.0 & 64 (64) \\
Code & 4B & 4,000 & $2\mathrm{e}{-5}$ & 0.03 & 0.0 & 0.2 & 64 (64) \\
Code & 8B & 2,000 & $5\mathrm{e}{-6}$ & 0.05 & 0.01 & 1.0 & 64 (32 $\times$ 2 accum.) \\
\midrule
Science & 0.6B & 2,000 & $3\mathrm{e}{-5}$ & 0.05 & 0.01 & 1.0 & 64 (64) \\
Science & 4B & 4,000 & $2\mathrm{e}{-5}$ & 0.03 & 0.0 & 0.2 & 64 (64) \\
Science & 8B & 2,000 & $5\mathrm{e}{-6}$ & 0.05 & 0.01 & 1.0 & 64 (32 $\times$ 2 accum.) \\
\bottomrule
\end{tabular}
\end{adjustbox}
\end{table}

All runs use AdamW with $\beta_1=0.9$ and $\beta_2=0.999$, cosine decay with decay ratio 0.9, minimum learning-rate ratio 0.1, maximum sequence length 32,768, RoPE $\theta=10^6$, seed 42, and the Qwen3 chat template. All experiments were run on Google Cloud preemptible TPU slices; representative wall-clock times ranged from roughly 5 hours for Qwen3-0.6B, about 49 hours for Qwen3-4B, and roughly 67 hours for Qwen3-8B.

\subsubsection{Evaluation}
\label{app:evaluations}

We evaluate \method{} across math, science, and coding benchmarks. Unless otherwise stated, we report pass@1. For benchmarks evaluated with multiple random seeds, we report average pass@1 across seeds. We use validation benchmarks for checkpoint selection and report final results only on held-out test benchmarks.

\paragraph{Math.}
We evaluate mathematical reasoning using both broad competition-math benchmarks and recent contest problems.
\begin{itemize}
    \item \textbf{MATH500} is a 500-problem subset of the MATH benchmark~\citep{lightman2023lets, hendrycks2021measuring}. MATH consists of competition-style high-school mathematics problems with final answers and step-by-step solutions, spanning topics such as algebra, geometry, number theory, probability, and precalculus. We use MATH500 as a validation benchmark for broad mathematical problem solving.
    \item \textbf{OlympiadBench Math} is the English-only math subset of OlympiadBench~\citep{he2024olympiadbenchchallengingbenchmarkpromoting}, an olympiad-level bilingual multimodal benchmark containing mathematics and physics problems from olympiad-style competitions and the Chinese college entrance exam. We use it as a validation benchmark for harder olympiad-style mathematical reasoning.
    \item \textbf{AIME24/AIME25/AIME26} are drawn from the American Invitational Mathematics Examination (AIME), a 15-question, 3-hour invitational competition administered by the Mathematical Association of America in which answers are integers from 0 to 999~\citep{maa_aime}. We use AIME24 and AIME25 for validation and AIME26 as a held-out test benchmark, evaluating each with 10 random seeds.
    \item \textbf{HMMT 02/25} is drawn from the Harvard-MIT Mathematics Tournament February 2025 archive~\citep{hmmt_feb_2025}. HMMT problems are recent competition-math problems spanning areas such as algebra, number theory, combinatorics, and geometry. We use HMMT 02/25 as a held-out test benchmark and evaluate with 10 random seeds.
\end{itemize}

\paragraph{Science.}
We evaluate scientific reasoning using benchmarks that test advanced STEM knowledge, multi-step problem solving, and expert-level question answering.
\begin{itemize}
    \item \textbf{JEEBench} consists of 515 challenging mathematics, physics, and chemistry problems curated from the IIT JEE-Advanced exam~\citep{arora2023llmsadvancedenoughchallenging}. These problems require long-horizon reasoning together with pre-engineering domain knowledge, and prior evaluations found that even strong models remain far from saturating the benchmark. We use JEEBench as a validation benchmark and evaluate with 3 random seeds.
    \item \textbf{OlympiadBench Physics} is the English-only physics subset of OlympiadBench~\citep{he2024olympiadbenchchallengingbenchmarkpromoting}. We use it as a validation benchmark for difficult physics reasoning.
    \item \textbf{GPQA Diamond} is the hardest commonly used subset of GPQA, a graduate-level multiple-choice benchmark written by domain experts in biology, chemistry, and physics~\citep{rein2023gpqagraduatelevelgoogleproofqa}. GPQA is designed to be ``Google-proof'': skilled non-experts with web access perform poorly, while domain experts do substantially better. We use GPQA Diamond as a held-out test benchmark and evaluate with 3 random seeds.
    \item \textbf{HLE} refers to Humanity's Last Exam, an expert-level closed-ended academic benchmark with broad subject coverage across mathematics, humanities, and the natural sciences~\citep{phan2025lastexam}. HLE questions are designed to be precise, verifiable, and resistant to simple internet lookup. We use HLE as a held-out test benchmark and evaluate with 3 random seeds.
\end{itemize}

\paragraph{Coding.}
We evaluate coding ability using temporally separated LiveCodeBench splits~\citep{jain2024livecodebenchholisticcontaminationfree}. LiveCodeBench continuously collects programming problems from LeetCode, AtCoder, and Codeforces, annotates problems by release date, and supports time-windowed evaluation to reduce contamination risk. We use the code-generation setting and report pass@1.
\begin{itemize}
    \item \textbf{LiveCodeBench v2} covers problems released from 5/2023--5/2024 and is used for validation. We evaluate with 6 random seeds.
    \item \textbf{LiveCodeBench v5 Official} covers problems released from 8/2024--1/2025 and is used as a held-out test benchmark. We evaluate with 6 random seeds.
    \item \textbf{LiveCodeBench v6 Official} covers problems released from 1/2025--4/2025 and is used as a held-out test benchmark. We evaluate with 6 random seeds.
\end{itemize}

\paragraph{Checkpoint selection.}
For each training run, we select checkpoints using validation benchmarks from the corresponding domain. Math checkpoints are selected using MATH500, OlympiadBench Math, AIME24, and AIME25. Science checkpoints are selected using JEEBench and OlympiadBench Physics. Coding checkpoints are selected using LiveCodeBench v2. After checkpoint selection, we report final results on the held-out test benchmarks: AIME26 and HMMT 02/25 for math, GPQA Diamond and HLE for science, and LiveCodeBench v5/v6 Official for coding.

\subsection{Number of Generations and Verification Strength}
\label{app:number_generations_filter_strength}

Tables~\ref{tab:num_generations_filter_strength_math}, \ref{tab:num_generations_filter_strength_science}, and \ref{tab:num_generations_filter_strength_code} report the full ablation results over the number of candidate generations $n$ and verification strength $v$.
Here, $n$ controls how many solutions are sampled for each unlabeled seed question, while $v$ controls how many repeated judge calls are used per verification stage. For each setting, we construct a self-curated dataset, train Qwen3-4B with the same SFT recipe, and report both validation and held-out test performance.

\begin{table}[h]
\caption{Math results from training Qwen3-4B on self-curated data constructed with varying generation and verification. $\Delta$ denotes the summed improvement over the initial Qwen3-4B model within each split.}
\label{tab:num_generations_filter_strength_math}
\centering
\small
\begin{adjustbox}{max width=\linewidth}
\begin{tabular}{lcccccccc}
\toprule
\textbf{Models}
& \multicolumn{5}{c}{\textbf{Validation}}
& \multicolumn{3}{c}{\textbf{Test}} \\
\cmidrule(lr){2-6} \cmidrule(lr){7-9}
& \textbf{MATH500} & \textbf{OlympiadBench} & \textbf{AIME24} & \textbf{AIME25} & \textbf{$\Delta$}
& \textbf{AIME26} & \textbf{HMMT} & \textbf{$\Delta$} \\
\midrule
Qwen3-4B & 88.4 & 61.7 & 65.7 & 58.7 & -- & 59.3 & 39.3 & -- \\
\midrule
No filter & 91.0 \positive{+2.6} & 63.8 \positive{+2.1} & 74.0 \positive{+8.3} & 61.0 \positive{+2.3} & +15.3 & 62.3 \positive{+3.0} & 41.0 \positive{+1.7} & +4.7 \\
$n=1$, $v=1$ & 90.0 \positive{+1.6} & 65.0 \positive{+3.3} & 73.3 \positive{+7.6} & 63.7 \positive{+5.0} & +17.5 & 63.7 \positive{+4.4} & 41.3 \positive{+2.0} & +6.4 \\
$n=1$, $v=3$ & 90.2 \positive{+1.8} & 63.2 \positive{+1.5} & 68.7 \positive{+3.0} & 64.3 \positive{+5.6} & +11.9 & 63.7 \positive{+4.4} & 42.0 \positive{+2.7} & +7.1 \\
$n=1$, $v=5$ & 90.4 \positive{+2.0} & 63.1 \positive{+1.4} & 72.0 \positive{+6.3} & 60.0 \positive{+1.3} & +11.0 & 60.3 \positive{+1.0} & 41.3 \positive{+2.0} & +3.0 \\
$n=4$, $v=1$ & 90.6 \positive{+2.2} & 64.2 \positive{+2.5} & 73.7 \positive{+8.0} & 62.0 \positive{+3.3} & +16.0 & 63.7 \positive{+4.4} & 41.3 \positive{+2.0} & +6.4 \\
$n=4$, $v=3$ & \textbf{91.6} \positive{+3.2} & 63.8 \positive{+2.1} & 73.3 \positive{+7.6} & 61.3 \positive{+2.6} & +15.5 & 64.0 \positive{+4.7} & 43.3 \positive{+4.0} & +8.7 \\
$n=4$, $v=5$ & 90.0 \positive{+1.6} & 65.0 \positive{+3.3} & 72.7 \positive{+7.0} & 62.3 \positive{+3.6} & +15.5 & 64.0 \positive{+4.7} & 42.0 \positive{+2.7} & +7.4 \\
$n=8$, $v=1$ & 89.8 \positive{+1.4} & \textbf{65.6} \positive{+3.9} & \textbf{76.0} \positive{+10.3} & 59.7 \positive{+1.0} & +16.6 & 61.0 \positive{+1.7} & 44.7 \positive{+5.4} & +7.1 \\
$n=8$, $v=3$ & 91.4 \positive{+3.0} & \textbf{65.6} \positive{+3.9} & 73.0 \positive{+7.3} & 62.3 \positive{+3.6} & +17.8 & 61.0 \positive{+1.7} & 45.7 \positive{+6.4} & +8.1 \\
$n=8$, $v=5$ & 90.6 \positive{+2.2} & 65.1 \positive{+3.4} & 73.0 \positive{+7.3} & \textbf{65.7} \positive{+7.0} & \textbf{+19.9} & \textbf{69.3} \positive{+10.0} & \textbf{46.0} \positive{+6.7} & \textbf{+16.7} \\
\bottomrule
\end{tabular}
\end{adjustbox}
\end{table}

\begin{table}[h]
\caption{Science results from training Qwen3-4B on self-curated data constructed with varying generation and verification.}
\label{tab:num_generations_filter_strength_science}
\centering
\small
\begin{adjustbox}{max width=\linewidth}
\begin{tabular}{lcccccc}
\toprule
\textbf{Models}
& \multicolumn{3}{c}{\textbf{Validation}}
& \multicolumn{3}{c}{\textbf{Test}} \\
\cmidrule(lr){2-4} \cmidrule(lr){5-7}
& \textbf{JEEBench} & \textbf{OlympiadBench Physics} & \textbf{$\Delta$}
& \textbf{GPQA Diamond} & \textbf{HLE} & \textbf{$\Delta$} \\
\midrule
Qwen3-4B & 66.4 & 13.3 & -- & 50.8 & 8.5 & -- \\
\midrule
No filter & 70.5 \positive{+4.1} & 16.9 \positive{+3.6} & +7.7 & 56.1 \positive{+5.3} & 9.7 \positive{+1.2} & +6.5 \\
$n=1$, $v=1$ & 70.3 \positive{+3.9} & 16.4 \positive{+3.1} & +7.0 & 55.1 \positive{+4.3} & 11.6 \positive{+3.1} & +7.4 \\
$n=1$, $v=3$ & 69.1 \positive{+2.7} & 17.5 \positive{+4.2} & +6.9 & 56.9 \positive{+6.1} & 10.0 \positive{+1.5} & +7.6 \\
$n=1$, $v=5$ & 69.3 \positive{+2.9} & 15.7 \positive{+2.4} & +5.3 & 55.1 \positive{+4.3} & 9.4 \positive{+0.9} & +5.2 \\
$n=4$, $v=1$ & \textbf{73.9} \positive{+7.5} & 16.5 \positive{+3.2} & \textbf{+10.7} & 54.7 \positive{+3.9} & 9.2 \positive{+0.7} & +4.6 \\
$n=4$, $v=3$ & 70.9 \positive{+4.5} & 17.7 \positive{+4.4} & +8.9 & 57.7 \positive{+6.9} & \textbf{14.7} \positive{+6.2} & \textbf{+13.1} \\
$n=4$, $v=5$ & 71.8 \positive{+5.4} & 16.7 \positive{+3.4} & +8.8 & 54.7 \positive{+3.9} & 10.3 \positive{+1.8} & +5.7 \\
$n=8$, $v=1$ & 70.7 \positive{+4.3} & 16.9 \positive{+3.6} & +7.9 & 54.5 \positive{+3.7} & 10.1 \positive{+1.6} & +5.3 \\
$n=8$, $v=3$ & 72.0 \positive{+5.6} & 17.1 \positive{+3.8} & +9.4 & 53.9 \positive{+3.1} & 10.9 \positive{+2.4} & +5.5 \\
$n=8$, $v=5$ & 70.3 \positive{+3.9} & \textbf{18.1} \positive{+4.8} & +8.7 & \textbf{60.4} \positive{+9.6} & 10.0 \positive{+1.5} & +11.1 \\
\bottomrule
\end{tabular}
\end{adjustbox}
\end{table}

\begin{table}[h]
\caption{Code results from training Qwen3-4B on self-curated data constructed with varying generation and verification.}
\label{tab:num_generations_filter_strength_code}
\centering
\small
\begin{adjustbox}{max width=\linewidth}
\begin{tabular}{lccccc}
\toprule
\textbf{Models}
& \multicolumn{2}{c}{\textbf{Validation}}
& \multicolumn{3}{c}{\textbf{Test}} \\
\cmidrule(lr){2-3} \cmidrule(lr){4-6}
& \textbf{LiveCodeBench} & \textbf{$\Delta$}
& \textbf{LCBv5} & \textbf{LCBv6} & \textbf{$\Delta$} \\
\midrule
Qwen3-4B & 70.6 & -- & 45.1 & 37.5 & -- \\
\midrule
No filter & 68.7 \negative{-1.9} & -1.9 & 43.0 \negative{-2.1} & 36.4 \negative{-1.1} & -3.2 \\
$n=1$, $v=1$ & 71.9 \positive{+1.3} & +1.3 & 46.4 \positive{+1.3} & 39.4 \positive{+1.9} & +3.2 \\
$n=1$, $v=3$ & 73.7 \positive{+3.1} & +3.1 & \textbf{49.3} \positive{+4.2} & 41.4 \positive{+3.9} & +8.1 \\
$n=1$, $v=5$ & 72.7 \positive{+2.1} & +2.1 & 48.9 \positive{+3.8} & 41.6 \positive{+4.1} & +7.9 \\
$n=4$, $v=1$ & 68.6 \negative{-2.0} & -2.0 & 44.8 \negative{-0.3} & 35.2 \negative{-2.3} & -2.6 \\
$n=4$, $v=3$ & 72.5 \positive{+1.9} & +1.9 & 47.1 \positive{+2.0} & 37.5 \positive{+0.0} & +2.0 \\
$n=4$, $v=5$ & 71.3 \positive{+0.7} & +0.7 & 47.4 \positive{+2.3} & 39.7 \positive{+2.2} & +4.5 \\
$n=8$, $v=1$ & 71.0 \positive{+0.4} & +0.4 & 46.2 \positive{+1.1} & 39.2 \positive{+1.7} & +2.8 \\
$n=8$, $v=3$ & 72.2 \positive{+1.6} & +1.6 & 45.5 \positive{+0.4} & \textbf{44.2} \positive{+6.7} & +7.1 \\
$n=8$, $v=5$ & \textbf{76.4} \positive{+5.8} & \textbf{+5.8} & 49.0 \positive{+3.9} & 41.9 \positive{+4.4} & \textbf{+8.3} \\
\bottomrule
\end{tabular}
\end{adjustbox}
\end{table}

\subsection{Alternative Verification}
\label{app:alternative_verification}

For the ablation comparing different verification strategies in Section~\ref{sec:type_of_verification}, we replace the UQ-style multi-stage verifier with a simpler prompt-based verifier. This verifier directly judges whether a generated solution is accurate and correct for the given question, without using the separate cycle-consistency, factuality, and correctness stages used by the full UQ-style verifier. The prompt is as follows:

\begin{Verbatim}
Please act as an impartial judge and evaluate whether the following answer is accurate and correct given the question.

Question: {question}
Answer: {answer}

Provide your decision: [[Y]] if the answer is accurate and correct, or [[N]] if it is not.
\end{Verbatim}

\subsection{Selection Policy}
\label{app:selection_policy}

Once verification is applied, a seed question may have zero, one, or multiple accepted candidate solutions. We therefore compare selection policies for converting verified candidates into the final training set. The default policy keeps the first accepted solution for each seed question, while the all-valid policy trains on every accepted solution.

Table~\ref{tab:selection_policy} shows that training on all accepted solutions is not always beneficial. Although the all-valid policy increases the amount of self-generated training data, it performs worse than the first-valid policy on both validation and held-out test benchmarks. This suggests that even verifier-accepted solutions are not equally useful: keeping all of them can introduce redundancy, overweight seed questions with many accepted candidates, or include lower-quality traces that passed verification. We therefore use the first-valid policy as our default selection rule.

\begin{table}[t]
\caption{Selection-policy ablation for Qwen3-4B math using $n=8$ generations and verification strength $v=5$. The default policy keeps a single accepted solution per seed question, whereas the all-valid policy trains on all accepted solutions. Val Avg. $\Delta$ is the average improvement over the initial model across MATH500, OlympiadBench, AIME24, and AIME25. Test Avg. $\Delta$ is the average improvement over the initial model across AIME26 and HMMT.}
\label{tab:selection_policy}
\centering
\small
\resizebox{\linewidth}{!}{%
\begin{tabular}{lccccc|ccc}
\toprule
& \multicolumn{5}{c|}{Validation} & \multicolumn{3}{c}{Test} \\
\cmidrule(lr){2-6} \cmidrule(lr){7-9}
Selection policy & MATH500 & OlympiadBench & AIME24 & AIME25 & Avg. $\Delta$ & AIME26 & HMMT & Avg. $\Delta$ \\
\midrule
First valid & \textbf{90.6} \positive{+2.2} & \textbf{65.1} \positive{+3.4} & \textbf{73.0} \positive{+7.3} & \textbf{65.7} \positive{+7.0} & \textbf{+5.0} & \textbf{69.3} \positive{+10.0} & \textbf{46.0} \positive{+6.7} & \textbf{+8.4} \\
All valid & 90.0 \positive{+1.6} & 64.1 \positive{+2.4} & 71.0 \positive{+5.3} & 60.3 \positive{+1.6} & +2.7 & 63.3 \positive{+4.0} & 45.0 \positive{+5.7} & +4.9 \\
\bottomrule
\end{tabular}%
}
\end{table}



\end{document}